%% file: main.tex
\documentclass[sigconf]{acmart}

\makeatletter
\def\@ACM@checkaffil{
    \if@ACM@instpresent\else
    \ClassWarningNoLine{\@classname}{No institution present for an affiliation}%
    \fi
    \if@ACM@citypresent\else
    \ClassWarningNoLine{\@classname}{No city present for an affiliation}%
    \fi
    \if@ACM@countrypresent\else
        \ClassWarningNoLine{\@classname}{No country present for an affiliation}%
    \fi
}
\makeatother

\usepackage{amsfonts}
\usepackage{amsmath}
\usepackage{booktabs}
\usepackage{threeparttable}
\usepackage{subfigure}
\usepackage{xspace,balance,tabularx,multirow}
\usepackage{flushend}
\usepackage{tikz}
\usepackage{pgfplots}

\usepackage{esvect}
\usepackage{enumitem}
\usepackage{url}

\usepackage{algorithmic}
\usepackage{algorithm}
\makeatletter
\newcommand{\removelatexerror}{\let\@latex@error\@gobble}
\makeatother

\include{cmd.tex}

\AtBeginDocument{%
  }

\setcopyright{acmlicensed}
\copyrightyear{2018}
\acmYear{2018}
\acmDOI{XXXXXXX.XXXXXXX}
\acmConference[CIKM '24]{International Conference on
Information and Knowledge Management}{October 21 –- 25, 2024}{Boise, Idaho, USA}
\acmISBN{978-1-4503-XXXX-X/18/06}

\begin{document}

\title{Effective Illicit Account Detection on Large Cryptocurrency MultiGraphs 
}

\author{Zhihao Ding}
\email{tommy-zh.ding@connect.polyu.hk}
\orcid{0000-0001-7778-6142}
\affiliation{%
  \institution{The Hong Kong Polytechnic University}
}

\author{Jieming Shi}
\authornote{Corresponding Author.}
\email{jieming.shi@polyu.edu.hk}
\orcid{0000-0002-0465-1551}
\affiliation{%
  \institution{The Hong Kong Polytechnic University}
}

\author{Qing Li}
\email{csqli@comp.polyu.edu.hk}
\orcid{0000-0003-3370-471X}
\affiliation{%
  \institution{The Hong Kong Polytechnic University}
}

\author{Jiannong Cao}
\email{csjcao@comp.polyu.edu.hk}
\orcid{0000-0002-2725-2529}
\affiliation{%
  \institution{The Hong Kong Polytechnic University}
}

\renewcommand{\shortauthors}{Ding et al.}

\begin{abstract}

Cryptocurrencies are rapidly expanding and becoming vital in digital financial markets. However, the rise in cryptocurrency-related illicit activities has led to significant losses for users. To protect the security of these platforms, it is critical to identify illicit accounts effectively. Current detection methods mainly depend on feature engineering or are inadequate to leverage the complex information within cryptocurrency transaction networks, resulting in suboptimal performance.
In this paper, we present \algo, an effective method for detecting illicit accounts in cryptocurrency transaction networks modeled by directed multi-graphs with attributed edges.
\algo first features an \edgeseq module that captures intrinsic transaction patterns from parallel edges by considering edge attributes and their directed sequences, to generate effective node representations.
Then in \algo, we design a multigraph Discrepancy (MGD) module with a tailored message passing mechanism to capture the discrepant features between normal and illicit nodes over   the multigraph topology,  assisted by an attention mechanism. 
\algo integrates these techniques for end-to-end training to detect illicit accounts from legitimate ones.
Extensive experiments, comparing against 15 existing solutions on 4 large cryptocurrency datasets of Bitcoin and Ethereum, demonstrate that \algo consistently outperforms others in accurately identifying illicit accounts. For example, on a Bitcoin dataset with 20 million nodes and 203 million edges, \algo attains an F1 score of 96.55\%, markedly surpassing the runner-up's score of 83.92\%.
The code is available at \url{https://github.com/TommyDzh/DIAM}.
\end{abstract}





\maketitle

\input{introduction.tex}

\input{relatedwork.tex}

\input{problem.tex}

\input{method.tex}

\input{experiments.tex}

\input{conclusion.tex}

\bibliographystyle{ACM-Reference-Format}
\bibliography{sample-base}

\end{document}

%% file: cmd.tex
\newcommand{\eat}[1]{{}}

\newcommand{\ie}{{\it i.e.},\xspace}
\newcommand{\eg}{{\it e.g.},\xspace}
\newcommand{\etal}{{\it et al.}\xspace}
\newcommand{\etc}{{\it etc.},\xspace}

\def\header{\vspace{1mm} \noindent}

\newcommand{\algo}{\textsf{DIAM}\xspace}

\newcommand{\edgeseq}{Edge2Seq\xspace}

\newcommand{\ethphish}{Ethereum-P\xspace}
\newcommand{\ethsecond}{Ethereum-S\xspace}
\newcommand{\bitcoinS}{Bitcoin-M\xspace}
\newcommand{\bitcoinL}{Bitcoin-L\xspace}
\newcommand{\pdetector}{Pdetector\xspace}
\newcommand{\sigtran}{SigTran\xspace}
\newcommand{\edgeprop}{EdgeProp\xspace}
\newcommand{\dci}{DCI\xspace}
\newcommand{\gdn}{GDN\xspace}
\newcommand{\caregnn}{CARE-GNN\xspace}
\newcommand{\transconv}{TransConv\xspace}
\newcommand{\gine}{GINE\xspace}
\newcommand{\gate}{GATE\xspace}
\newcommand{\gat}{GAT\xspace}
\newcommand{\sage}{Sage\xspace}
\newcommand{\gcn}{GCN\xspace}
\newcommand{\fraudre}{FRAUDRE\xspace}
\newcommand{\pc}{PC-GNN\xspace}
\newcommand{\bert}{BERT4ETH\xspace}

\def\eout{{e}}
\def\Ea{\mathbf{X}_E}
\def\YL{Y_\mathcal{L}\xspace}
\def\YU{Y_\mathcal{U}\xspace}
\def\yv{y_v}

\def\Evout{E^{out}_v}
\def\Eavout{X^{out}_v}
\def\Eavin{X^{in}_v}

\newcommand{\eavt}[1]{{\mathbf{x}_{e_v^#1}}}
\newcommand{\ezvt}[1]{{\mathbf{z}^{out}_{e_v^#1}}}

\newcommand{\houtvt}[1]{{\mathbf{h}^#1_{v_{out}}}}
\newcommand{\houtv}{{\mathbf{h}_{v_{out}}}}
\newcommand{\hinv}{{\mathbf{h}_{v_{in}}}}
\newcommand{\hv}{{\mathbf{h}_{v}}}
\newcommand{\ghvl}[1]{{\mathbf{h}^#1_{v}}}

\newcommand{\gzvl}[1]{{\mathbf{z}^#1_{v}}}
\newcommand{\gzvnol}{{\mathbf{z}_{v}}}
\newcommand{\gzul}[1]{{\mathbf{z}^#1_{u}}}
\newcommand{\gdvoutl}{{\mathbf{r}^{(\ell)}_{v_{out}}}}
\newcommand{\gdvinl}{{\mathbf{r}^{(\ell)}_{v_{in}}}}
\newcommand{\gdvoutnol}{{\mathbf{r}_{v_{out}}}}
\newcommand{\gdvinnol}{{\mathbf{r}_{v_{in}}}}
\newcommand{\Nin}[1]{N_{in}(#1)}
\newcommand{\Nout}[1]{N_{out}(#1)}
\newcommand{\awv}[1]{\alpha_{v,#1}}

%% file: introduction.tex
\section{Introduction}\label{sec:introduction}

Cryptocurrencies, \eg Ethereum and Bitcoin, are of growing importance, due to the nature of decentralization and pseudo-anonymity based on blockchain technology. 
As of May 2024, Bitcoin and Ethereum are the top-2 largest cryptocurrencies with \$1.5 trillion market capitalization in total \cite{coinmarketcap_2023}.
In addition to the cryptocurrency transactions among normal accounts, illicit accounts are also taking advantage of Bitcoin and Ethereum for illegal activities, such as phishing scams \cite{chen2020ijcai,chen2018detecting}, and money laundering \cite{wu2021detecting}, which put normal users at risk of financial loss and hinder the development of the blockchain ecosystem.

\eat{
\begin{table}[t!]
  \centering
  \caption{The percentage of normal nodes in the neighborhood of illicit nodes.}
  \resizebox{0.9\linewidth}{!}{ 
    \begin{tabular}{ccccc}
    \toprule
    Dataset  & {Ethereum-S} & Ethereum-P & {Bitcoin-M} & {Bitcoin-L} \\
    \midrule
          & 97.41 & 98.23 & 76.91 & 84.78 \\
    \bottomrule
    \end{tabular}}
  \label{tab:heterophily}
  \vspace{-6mm}
\end{table}
}

Hence, we study the detection of illicit accounts on cryptocurrency transaction networks. This task is particularly challenging due to the huge number of transactions and the inherent anonymity of cryptocurrency accounts, which lack the portrait information crucial for identifying illicit activities.
Some pioneer solutions  \cite{akcora2020bitcoinheist,chen2020ijcai,poursafaei2021sigtran,chen2018detecting} mainly rely on feature engineering to extract handcrafted features from transactions, which highly depends on domain expertise. 
There are also studies using Graph Neural Networks (GNNs) for detection \cite{wu2020phishers,poursafaei2021sigtran,weber2019anti,tam2019identifying,chen2020toit}.
Common GNNs, such as GCNs \cite{kipf2017semi} and  GATs \cite{velivckovic2017graph},  mainly rely on the homophily  assumption that connected nodes share similar representations  and belong to the same class \cite{zhu2020beyond}.
This may not be true for illicit account detection.
Specifically, illicit accounts are usually much fewer than normal accounts, and they may exhibit discrepant patterns over their neighboring accounts, most of which are normal. Such discrepancy should be captured for effective detection of illicit accounts.
A recent method~\cite{hu2023bert4eth} adopts transformers to learn account representations from Ethereum transaction sequences for detection, without explicitly considering the network topology of transactions.
As reviewed in Section \ref{sec:relatedwork}, general graph anomaly detection methods \cite{ding2021few,liu2021intention} can be customized for illicit account detection, but yield moderate performance in experiments.

\begin{figure}[!t]
	\centering
		\includegraphics[width=0.85\linewidth]{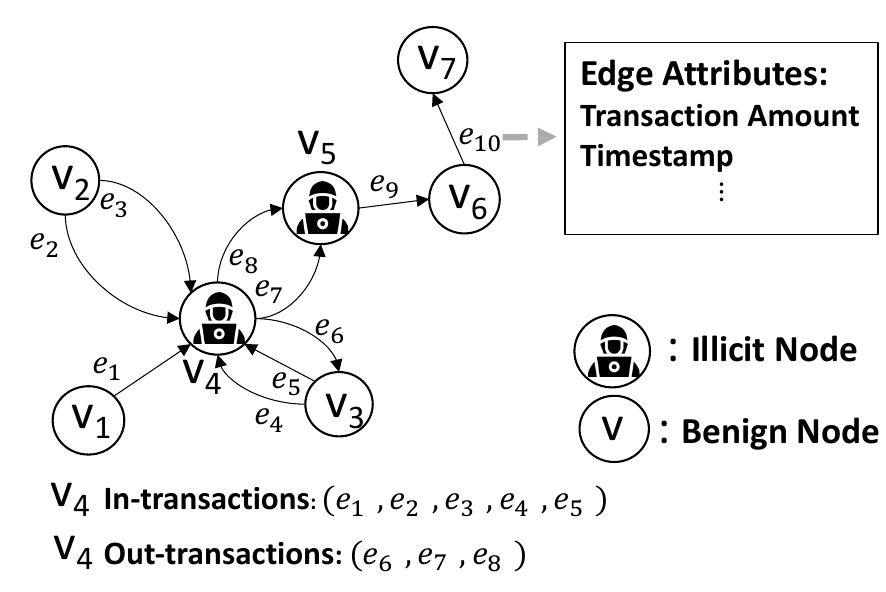}
  \vspace{-2mm}
 \caption{A directed multigraph with edge attributes. 
}
\label{fig:illus}
\vspace{-3mm}
\end{figure}

To effectively model the cryptocurrency transaction network, we conceptualize it as a \textit{directed multi-graph} with multiple edges connecting nodes, each edge representing a transaction between accounts. These edges are enriched with {edge attributes} such as transaction timestamps and amounts, enabling comprehensive representation of transactional activities in both Bitcoin and Ethereum networks. An illustrative example is depicted in Figure \ref{fig:illus}. For instance, edge $e_{10}$ is a transaction from nodes $v_{6}$ to $v_{7}$ with transaction timestamp, value, \etc as edge attributes. Multiple edges exist between nodes, \eg edges $e_4,e_5,e_6$ between $v_3$ and $v_4$, representing three transactions. 
A node, \eg $v_4$, has incoming and outgoing transactions as listed in Figure \ref{fig:illus}. The transaction timestamps indicate the sequential dependency of edges between nodes.

In this paper, we present \algo, an effective method  to \underline{D}etect \underline{I}llicit \underline{A}ccounts on directed \underline{M}ultigraphs with edge attributes for cryptocurrencies.
In a nutshell, \algo consists of well-thought-out technical designs to holistically utilize all of the directed multigraph topology, edge attributes, and parallel edge sequential dependencies, as shown in Figure \ref{fig:illus}.
First, \algo incorporates an \textit{\edgeseq} module designed to autonomously learn effective representations that maintain the inherent transaction patterns depicted by directed parallel edges with attributes. In particular, \edgeseq identifies and captures the sequential patterns of transactions by assembling sequences of edge attributes. It then integrates both the attributes of the edges and the dependencies within these sequences into the representations of nodes.
To further utilize the multigraph topology and handle the discrepancy issue mentioned above, we then develop an \textit{Multigraph Discrepancy} (\textit{MGD})  module in \algo. 
As illustrated in Figure \ref{fig:illus}, illicit node $v_4$ is closely connected to benign nodes $v_1,v_2,v_3$, while their representations should be discrepant to distinguish $v_4$ from others. To achieve this, we design MGD as a message-passing mechanism to propagate not only node representations, but also the discrepancies between nodes, along directed multiple edges, with the help of a dedicated attention mechanism and learnable transformation. In other words, MGD can preserve both similar and discrepant features, which are vital for effective illicit account detection. 
\algo stacks multiple MGD modules to consider multi-hop multigraph topology.
Finally, assembling all techniques, \algo is trained in an end-to-end manner, to minimize a cross-entropy loss.
We evaluate \algo against 15 existing solutions over 4 large real-world cryptocurrency datasets of Bitcoin and Ethereum. 
Extensive experiments validate that \algo consistently achieves the highest accuracy on all datasets, outperforming competitors often by a significant margin, while being efficient.

Summing up, our contributions  are as follows:
\vspace{1mm}
\begin{itemize}
  \item We study illicit account detection on transaction networks of cryptocurrencies, and present \algo, an effective method over large directed multigraphs with edge attributes.

  \item In \algo, we develop an \edgeseq module that automatically encodes edge attributes, edge sequence dependencies, and edge directions into node representations.
  \item We further design MGD, a multigraph discrepancy module to effectively preserve the representation discrepancies between illicit and benign nodes on the multigraph.
  \item The superiority of \algo is validated via extensive experiments by comparing 15 baselines on 4 real datasets.
\end{itemize}

%% file: relatedwork.tex
\section{Related Work} \label{sec:relatedwork}
Our work is related to studies on illicit account detection on cryptocurrency, and graph-based anomaly detection.

\header
\textbf{Illicit Account Detection on Cryptocurrency.}
Early studies  mostly rely on tedious feature engineering to obtain statistical features, such as the sum, average, standard deviation of transaction amounts and time \cite{chen2020toit,chen2020ijcai,poursafaei2021sigtran}.
These studies then employ on-the-rack classifiers (\eg XGBoost \cite{chen2016xgboost} and LightGBM \cite{ke2017lightgbm}) over the extracted features to detect illicit accounts \cite{akcora2020bitcoinheist, chen2020ijcai,chen2020toit}.
To further exploit the graph topological characteristics of cryptocurrency transaction networks, recent studies \cite{wu2020phishers, poursafaei2021sigtran}  incorporate graph mining methods  for illicit account detection. 
Random-walk based node embedding is adopted in \cite{wu2020phishers}, and Node2vec~\cite{grover2016node2vec} and Ri-walk~\cite{ma2019riwalk} are used in~\cite{poursafaei2021sigtran} to extract structural information for 
illicit account detection. 
Most of these studies still use on handcrafted node features. 
A recent method ~\cite{hu2023bert4eth} uses transformers to learn expressive representations from Ethereum transaction sequences, but does not exploit the multigraph structure of cryptocurrency transactions.
There are GNN-based methods on cryptocurrency transaction networks \cite{weber2019anti,tam2019identifying,chen2020toit,li2021self}. An end-to-end GCN is trained in \cite{weber2019anti} for anti-money laundering in Bitcoin. 
\edgeprop~\cite{tam2019identifying}  augments edge attributes in GNNs to identify  illicit accounts in Ethereum.  In \cite{li2021self}, GNNs and self-supervised learning are incorporated to detect phishing scams. These studies usually focus on one cryptocurrency type, either Ethereum or Bitcoin. 
On the other hand,  we exploit the topological and sequential semantics of the directed multigraph data model, and develop techniques to automatically learn deep intrinsic node representations that are highly effective for illicit account detection on both Bitcoin and Ethereum cryptocurrencies.

\header
\textbf{Graph-based Anomaly Detection.}
There exist studies on anomaly detection over general graphs \cite{zhao2020error,hu2019strategies,shi2020masked,dou2020enhancing,ding2021inductive,zhou2021subtractive,liu2021pick,ding2021few,wang2021decoupling,zhang2021fraudre,abs-2307-15244}, 
and representative methods are based on classic GNNs, such as \gcn \cite{kipf2017semi}, \sage \cite{hamilton2017inductive}, and \gat \cite{velivckovic2017graph}. 
GINE~\cite{hu2019strategies} and TransConv~\cite{shi2020masked}  incorporate edge features in GNNs for anomaly detection. 
However, abnormal nodes may  have {discrepant} features, compared with normal ones \cite{liu2020alleviating}, and often hide themselves in camouflage \cite{dou2020enhancing}. 
To alleviate the issue,  \caregnn~\cite{dou2020enhancing} trains a predictor to measure the similarity between target nodes and their neighborhoods and adopts  reinforcement learning for detection. In \cite{ding2021inductive}, a new   framework is proposed to use attention mechanism and generative adversarial learning.
Camouflage behaviors are captured by subtractive aggregation on GNNs in  \cite{zhou2021subtractive}.
PC-GNN~\cite{liu2021pick} samples neighbors from the same class and relieve the imbalance between abnormal and normal nodes, while meta-learning is used in~\cite{ding2021few} and decoupling with self-supervised learning is developed in ~\cite{wang2021decoupling}.
\fraudre \cite{zhang2021fraudre} takes mean aggregation of neighborhood differences and representations,  and develops a loss function to remedy class imbalance for anomaly detection.
Note that these methods  are designed for relation graphs, and we set the number of relations as 1 to run them on the multigraph data model in this work.
However, though these methods can be customized for the illicit account detection problem, they often produce suboptimal performance in experiments, since they are not catered for the unique characteristics of cryptocurrency transactions.
Contrarily, we consider all aspects  of the multigraph data model into \algo for illicit account detection.

%% file: problem.tex
\begin{figure*}[!t]
	\centering
		\includegraphics[width=0.98\textwidth]{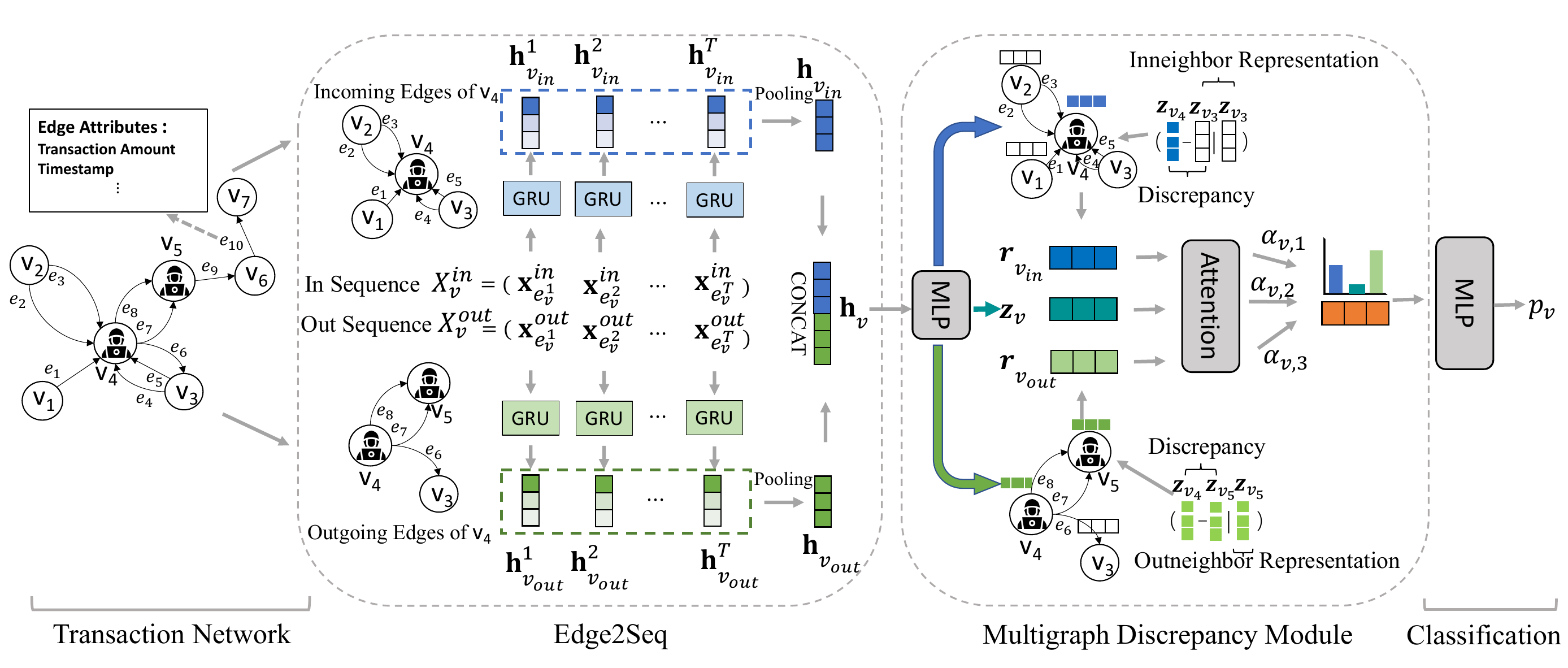}
  \vspace{-1mm}
 \caption{The \algo framework with an input transaction network modeled as a directed multigraph with edge attributes.}
\label{fig:frame}   
\vspace{-1mm}
\end{figure*}

\section{Problem Formulation}\label{sec:preliminary}

\textbf{Data Model.}
Let $G=(V,E,\Ea)$ be a directed multigraph, consisting of (i) a node set $V$ that contains $n$ nodes, (ii) a set of directed edges $E$ of size $m$, and (iii) an edge attribute matrix $\Ea\in\mathbb{R}^{m\times d}$, each row of which is a $d$-dimensional vector serving as the edge attributes to encode the details of the corresponding transaction.
In a multigraph $G$, nodes $v$ and $u$ can have parallel edges with different edge attributes.
Let $\Nout{v}$ be the \textit{multiset} of node $v$'s outgoing neighbors. If a node $u$ has more than one edge to $v$, $u$ will have multiple occurrences in $\Nout{v}$.
Similarly, let $\Nin{v}$ be the {multiset} of node $v$'s incoming neighbors.

Given a collection of transactions, we can build its directed edge-attributed multigraph as follows.  
An Ethereum transaction is a message sent from a sender address (i.e., account) $v$, to a receiver address $u$ at a certain time with transaction details, forming a directed edge,  \eg edge $e_{10}$ in Figure \ref{fig:illus}.
Bitcoin transactions are similar but with differences. A bitcoin transaction can contain multiple sender accounts and receiver accounts, who may send or receive different amounts of Bitcoin respectively in the transaction \cite{wu2021towards}.
Given a Bitcoin transaction, we will create a directed edge $e$ from every sender $v$ to every receiver $u$ in the transaction, with the corresponding transaction details from $v$ to $u$ as edge attributes.
For interested readers, see \cite{wu2021analysis} for a comprehensive introduction of Bitcoin and Ethereum.

\eat{\header
\textbf{Data model.}

Transactions can be treated as the interactions among accounts. 
Here we focus on how to build directed multigraphs with edge attributes using transaction data, and adopt Ethereum and Bitcoin transactions to explain.  
A transaction $e$  sent from  accounts $v$ to $u$ can be regarded as a directed edge from nodes $v$ to $u$ with edge attributes describing transaction details, such as transaction amount and timestamp (). Parallel edges may exist between $v$ and $u$, since there could be many transactions between nodes $v$ and $u$. 
Ethereum transactions follow the procedure above to model transactions into multigraphs. }

\header
\textbf{Problem Definition.} 
Given a directed multigraph $G=(V,E,\Ea)$
we formulate the problem of illicit account detection on directed multigraphs with edge attributes as a classification task. 
Let $\YL$ be the set of the partially observed node labels, and each node label $\yv\in\YL$ takes value either $1$ or $0$, indicating the node to be illicit or not. The objective  is to learn a binary classifier $f$ that can accurately detect the illicit accounts in the set of unobserved node labels $\YU$ to be predicted in $G$, $f:G=(V,E,\Ea,\YL) \mapsto \YL\cup\YU$.

Bitcoin and Ethereum are distributed 
public ledgers recording all transactions  anonymously accessible to the public  \cite{chen2020toit,zheng2017overview}, which facilitates the build of multigraphs.
For node labels, since the addresses in cryptocurrencies are unique and immutable, there are websites and forums, like WalletExplorer~\cite{walletexplorer_2022} and EtherScan~\cite{etherscan_2022}, providing illicit label information, e.g., phishing. As described in Section \ref{sec:expsetup}, we crawl such information as ground-truth labels.

%% file: method.tex
\section{The Proposed Method}\label{sec:method}

\header
\textbf{Overview.}
Figure \ref{fig:frame} illustrates the \algo method, which inputs a directed, edge-attributed multigraph $G$ representing a transaction network. 
The first module in \algo is {\edgeseq} detailed in Section \ref{sec:edge2seq}, which 
automatically derives the expressive representation of a node by considering the sequences of both incoming and outgoing edges.
As shown in Figure \ref{fig:frame}, for a node $v$ (\eg $v_4$), \edgeseq first builds an incoming sequence $\Eavin$ and an outgoing sequence $\Eavout$ that consist of $v$'s incoming and outgoing edge attributes in chronological order, respectively.
Intuitively, $\Eavout$ and $\Eavin$ describe different sequential transaction patterns  of node $v$, when $v$ serves as a sender or a receiver respectively.
Then \edgeseq employs GRUs ~\cite{cho2014properties} 
to learn the sequence representations of both $\Eavout$ and $\Eavin$, which are then processed by  pooling operations, to get representations $\houtv$ and $\hinv$ respectively.
Then $\houtv$ and $\hinv$ are concatenated together to be the node representation $\hv$ of $v$,  encapsulating the bidirectional transaction patterns and their temporal dependencies.
The node representations $\hv$ for all $v\in V$ learned by \edgeseq are then regarded as initial inputs fed into the proposed multigraph discrepancy (MGD) module presented in Section \ref{sec:gdl}.
In an MGD, a target node $v$  receives messages from its incoming and outgoing neighborhoods separately (\eg $v_4$  in the multigraph discrepancy module of Figure \ref{fig:frame}). The incoming and outgoing messages, denoted as $\gdvinnol$ and $\gdvoutnol$, contain {both} neighbor representations and their \textit{discrepancies} with the target node, in order to preserve distinguishable features for illicit account detection.
Then an attention mechanism is designed in MGD to integrate $v$'s representation $\gzvnol$, incoming  message $\gdvinnol$,  and outgoing message $\gdvoutnol$ together via attentions $\awv{1}$, $\awv{2}$, and $\awv{3}$. 
\algo stacks multiple MGD layers to   consider  multi-hop multigraph topology to learn more expressive discrepancy-aware node representations. 
The last component of \algo is a multilayer perceptron (MLP) to learn illicit probability $p_v$ of node $v$.
\algo is trained to minimize a binary cross-entropy loss in Section \ref{sec:obj}.

\subsection{\edgeseq}
\label{sec:edge2seq}

High-quality node representations are essential for detecting illicit accounts. As discussed in Section \ref{sec:introduction}, cryptocurrency accounts often lack profile information, and
illicit accounts in transaction networks frequently disguise their native features to blend in with legitimate nodes, a challenge exacerbated by the decentralized and pseudo-anonymous nature of cryptocurrencies. Current methods mostly rely on feature engineering to extract statistical features, which demands domain expert knowledge.

Here we develop \edgeseq to automatically generate high-quality node representations that capture the essential transaction patterns within nodes. Briefly, \edgeseq combines edge attributes (transaction details), parallel edge sequential dependencies (transaction relationships), and edge directions (transaction flow directions) within the directed edge-attributed multigraph data model.
In particular, \edgeseq treats the incoming and outgoing edges of a node distinctly, recognizing that they represent different directions of money flow, which is key for identifying transaction patterns in cryptocurrency networks. 
To effectively discern these directional transaction patterns, our model constructs separate incoming and outgoing sequences for each node $v$ in multigraph $G$, sorted by timestamps. We employ GRUs to process these sequences to learn representations. These representations then serve as the node representations for further training.
Next, we detail \edgeseq in two main steps: edge sequence generation and edge sequence encoding.

\eat{Obtaining high-quality node representations is crucial for illicit account detection  task. 
However, as mentioned in Section \ref{sec:introduction}, the native node features of illicit accounts are often falsified or lacking  in transaction networks, since these accounts intend to pretend themselves to be benign and hide themselves among normal nodes, particularly on cryptocurrencies that are decentralized and pseudo-anonymous \cite{chen2020toit}.
Existing solutions mostly resort to manual feature engineering to get statistical features \cite{chen2020ijcai,poursafaei2021sigtran}, which  requires domain expertise and is   dependent on a specific cryptocurrency. }

\eat{Here we develop \edgeseq to automatically learn high-quality node representations that preserve the   intrinsic transaction patterns of nodes. 
In a nutshell, \edgeseq integrates (i) {edge attributes} (transaction information), (ii) {parallel edge sequential dependencies} (transaction dependencies), and  (iii) {edge directions} (directional transaction flows) together in the multigraph data model.}

\eat{Remark that \edgeseq handles the incoming and outgoing edges of a node separately.
In fact, the incoming and outgoing edges of a node indicate  different money flow directions, whose differences are crucial to distinguish transaction patterns in cryptocurrency transaction networks \cite{tam2019identifying}.
For example, Chen \etal~\cite{chen2020toit} find that phishing accounts in Ethereum usually have fewer incoming edges, while having more outgoing edges, compared with non-phishing accounts. In addition, phishing accounts often receive 3 times more cryptocurrency amount than the amount spent. }

\eat{Hence, a model should be able to differentiate and capture such directional transaction patterns.
Specifically, for every node $v$ of the input multigraph $G$, we first build an incoming (resp. outgoing) sequence that consists of its incoming (resp. outgoing) edge attributes ordered by timestamps.
We then apply GRUs over the sequences to learn representations that preserve both edge attributes (\ie transaction information) and sequential dependencies among edges (\ie transaction behaviors). The sequence representations are then  processed as the respective node representations for subsequent training. 
In the following, we explain the details of \edgeseq that contains two parts: sequence generation and sequence encoding.}

\header
\textbf{Edge Sequence Generation.}
Given a node $v$ of the input multigraph $G$, \edgeseq first builds two sequences for it.
In particular, for all outgoing edges of $v$, \edgeseq sorts the outgoing edges in chronological order according to the timestamps on edges, and gets $\Evout=(\eout^1_{v},\eout^2_{v}..., \eout^T_{v})$, the sequence of $T$ sorted outgoing edges of $v$. 
For instance, in Figure \ref{fig:illus}, node $v_4$ has outgoing edge sequence $(e_6, e_7,e_8)$.
\edgeseq then extracts the corresponding edge attributes accordingly, and builds the outgoing edge attribute sequence of $v$, $\Eavout=(\eavt{1},\eavt{2},..., \eavt{T})$.
Then, similarly, we also build an incoming edge attribute sequence $\Eavin$.
Obviously, sequences $\Eavout$ and $\Eavin$ of node $v$ consider both edge sequence and edge attributes, and also utilize parallel edges between $v$ and its neighbors.
Intuitively, $\Eavout$ (resp. $\Eavin$) represents the transaction behaviors of node $v$ when $v$ serves as a sender (resp. receiver). 

Note that an account can participate in thousands of transactions, resulting to substantially long sequences. 
The number of transactions  of accounts commonly follows the power-law distribution \cite{chen2020understanding}. In other words, only a few nodes  have excessively  long sequences $\Eavout$ or $\Eavin$.
To reduce the computational costs associated with handling extremely long sequences, we apply a common trick \cite{vaswani2017attention,liu2021intention} by limiting the sequence length to be at most $T_{\max}$ and retaining the most recent edges. In experiments, we study the impact of varying $T_{\max}$.
In addition, for nodes without any incoming or outgoing edges, we add self-loops to generate sequences.

\header
\textbf{Edge Sequence Encoding.}
After generating sequences $\Eavout$ and $\Eavin$ for  node $v$ in the input multigraph $G$, we encode the sequences into the representation of node $v$.
We use node $v$'s length-$T$ outgoing sequence $\Eavout=(\eavt{1},\eavt{2},..., \eavt{T})$ to explain the process, and that of $\Eavin$ naturally follows.
In particular, as shown in Eq. \eqref{eq:seqEncode},  starting from $t=1$, until the end of the length-$T$ sequence $\Eavout$, we first apply a linear transformation on edge attributes $\eavt{t}$ to get $\ezvt{t}$ via a one-layer MLP with learnable $\mathbf{W}_{out}$ and $\mathbf{b}_{out}$.
Then we apply GRU over $\ezvt{t}$ and the $(t-1)$-th hidden state $\houtvt{{t-1}}$, to get the updated  $\houtvt{{t}}$ at the $t$-th position of sequence $\Eavout$:
\begin{equation} \label{eq:seqEncode}
\begin{split}
&\ezvt{t}=\mathbf{W}_{out}\eavt{t}+\mathbf{b}_{out},\\
&\houtvt{t}=\mathrm{GRU}_{out}(\ezvt{t},\houtvt{{t-1}}),
\end{split}
\end{equation}
where $\mathbf{W}_{out}\in\mathbb{R}^{\frac{c}{2}\times d}$ and $\mathbf{b}_{out}\in\mathbb{R}^{\frac{c}{2}}$ are learnable parameters, and $c$ is the representation dimension. By convention, the initial hidden state of GRU, $\houtvt{{t=0}}$, is set to be zero.

Essentially, we   generate a representation $\houtvt{t}$ for each outgoing edge at position $t\in[1,T]$ of  sequence $\Eavout$. 
Then we apply element-wise max-pooling  to get the representation $\houtv$ of sequence $\Eavout$, 
\begin{equation} \label{eq:maxpooling}
{\houtv = {\varphi_{pool}}_{\forall t\in[1,T]}(\houtvt{t}),}
\end{equation}
{where $\varphi_{pool}(\cdot)$ is the  max-pooling operation.}

Then, we apply a similar procedure over the incoming sequence $\Eavin$ of node $v$ by using another $\mathrm{GRU}_{in}$, to get the incoming sequence representation $\hinv$.
Finally, we obtain the  representation $\hv$ of node $v$ by concatenating  $\hinv$ and $\houtv$ in Eq. \eqref{eq:concat}. 
\begin{equation} \label{eq:concat}
\hv=\houtv||\hinv.
\end{equation}

Since we obtain the representations   $\hinv$ and $\houtv$ based on the incoming and outgoing edge attribute sequences of $v$ respectively, inherently node representation $\hv$ can preserve the hidden transaction patterns of node $v$ in both directions.

\input{MGDLayer}

\subsection{Objective}
\label{sec:obj}
\algo works in an end-to-end manner to detect illicit accounts on directed multigraphs with edge attributes.
At the last $L$-th MGD layer of \algo, we get the final representations $\ghvl{{(L)}}$ of nodes $v$.
For all labeled nodes $v$, we send their representations into a binary classifier, which is a 2-layer MLP network with a sigmoid unit as shown in Eq. \eqref{eq:classifier}, to generate the illicit probability $p_v$ of a node $v$. Obviously, $1-p_v$ is the normal probability of node $v$. 
\begin{equation} \label{eq:classifier}
p_{v} = \mathrm{sigmoid}(\mathrm{MLP}(\ghvl{{(L)}}))
\end{equation}

We adopt the standard binary cross-entropy loss for training:
\begin{equation} \label{eq:loss}
\mathrm{Loss}(\boldsymbol{\Theta}) = -\sum_{y_{v}\in \YL}(y_{v}\log(p_{v})+(1-y_{v})\log(1-p_{v})),
\end{equation}
where $\YL$ is the set of groundtruth node labels, $y_{v}$ is the label of node $v$,  $\boldsymbol{\Theta}$ contains all parameters of  \algo.

\header 
\textbf{Analysis.} We provide the time complexity analysis of \algo.
In \edgeseq, the time complexities of one-layer MLP transformation, GRU, max-pooling are $\mathcal{O}(T_{\max}|V|dc)$, $\mathcal{O}(T_{\max}|V|c^2)$, and $\mathcal{O}(T_{\max}|V|c)$ respectively, where $T_{\max}$ is the maximum sequence length, $|V|$ is the number of nodes, $d$ and $c$ are the dimensions of edge attributes and hidden representations. The overall time complexity of \edgeseq is $\mathcal{O}(T_{\max}|V|c(c+d))$. In MGD, the time complexity of message passing operation on incoming and outgoing neighbors is the same as vanilla message passing-based GNNs like \sage~\cite{hamilton2017inductive} and \gat~\cite{velivckovic2017graph}, which is $\mathcal{O}(|V|c^2+|E|c)$,  where $|E|$ is the number of edges. The time complexity of attention mechanism is $\mathcal{O}(|V|c)$, and the time complexity of the two-layer MLP is $\mathcal{O}(|V|(c^2+2c))$.
Combining the time of all above components, we get the time complexity of \algo as 
$\mathcal{O}(T_{\max}|V|c(c+d)+|E|c)$.

%% file: MGDLayer.tex
\subsection{MGD}
\label{sec:gdl}
Note that the representation $\hv$ of node $v$ obtained by \edgeseq in Section \ref{sec:edge2seq} only captures $v$'s individual transaction features contained in its outgoing and incoming edges.
In this section, we aim to leverage the multi-hop multigraph topology to enhance the representation for illicit account detection. One straightforward way is to adopt conventional GNNs.
However, as explained, conventional GNNs heavily rely on the  assumption that similar nodes tend to connect to each other and share similar representations \cite{hamilton2017inductive}, which may be less effective on the task of illicit account detection on multigraphs \cite{dou2020enhancing,ding2021inductive}.
Intuitively, illicit and normal nodes, despite potential close connections, should have distinct representations. An effective model should be capable of learning these discrepant representations between closely connected normal and illicit nodes.

To accomplish  this, we present a Multigraph Discrepancy module (MGD), with  three technical designs: (i) directed discrepancy-aware message passing with sum pooling, (ii) layer-wise learnable  transformations, and (iii) an attention mechanism over directional representations, to learn expressive representations.

The MGD is discrepancy-aware, transforming and transmitting not just node representations, but also the discrepancies between nodes through a proposed message passing mechanism on multigraphs. Moreover, for a target node $v$, MGD separately considers the discrepancies of its incoming and outgoing neighbors, acknowledging that a node's behavior can vary as either a sender or receiver of transactions. As confirmed in our experiments, MGD outperforms existing counterparts in learning distinct representations for illicit account detection.

In  \algo, let $L$ be the total number of MGD modules stacked together.
The first MGD layer takes the  representations $\hv$ of nodes $v\in V$ learned by \edgeseq in Section \ref{sec:edge2seq} as input.
Without ambiguity, let $\hv^{(\ell=0)}$ represent the input of the first MGD layer.
As shown in Eq. \eqref{eq:gdl},
the $\ell$-th MGD first applies a layer-wise linear transformation with learnable weights $\mathbf{W}_{2}^{(\ell)}$ and $\mathbf{b}_{2}^{(\ell)}$ to convert  representation $\ghvl{{(\ell-1)}}$ to intermediate $\gzvl{{(\ell)}}$ via a one-layer MLP. 
Then for an in-neighbor $u\in\Nin{v}$, the message passed from $u$ to $v$ in the $\ell$-th MGD is $\mathbf{W}_{3}^{(\ell)}(\gzul{{(\ell)}}||(\gzvl{{(\ell)}}-\gzul{{(\ell)}}))$, which includes both in-neighbor $u$'s representation  $\gzul{{(\ell)}}$ and its discrepancy $(\gzvl{{(\ell)}}-\gzul{{(\ell)}})$ with target node $v$, followed by a learnable linear transformation using $\mathbf{W}_{3}^{(\ell)}$. 
Aggregating all such information for every $u\in\Nin{v}$, we obtain $\gdvinl$ that is the  {discrepancy-aware incoming message} that node $v$ receives from its incoming neighborhood.
Note that $\Nin{v}$ is a multiset of node $v$'s in-neighbors in the input multigraph $G$, and thus, we consider parallel edges during the message passing.
Similarly, we can get the {discrepancy-aware outgoing message} $\gdvoutl$ that $v$ receives from its outgoing neighborhood $\Nout{v}$, as shown in Eq. \eqref{eq:gdl}. Specifically, $\gdvoutl$ considers every out-neighbor $u$'s representation as well as its discrepancy with $v$.
Finally, we develop an attention mechanism to integrate the three aspects, namely $v$'s representation $\gzvl{{(\ell)}}$, discrepancy-aware incoming and outgoing messages $\gdvinl$ and $\gdvoutl$, via attention $\awv{1}$, $\awv{2}$, and $\awv{3}$, to get node representation $\ghvl{{(\ell)}}$ at the $\ell$-th MGD. 

\begin{equation} \label{eq:gdl}
\begin{split}
&\gzvl{{(\ell)}}=\mathbf{W}_{2}^{(\ell)}\ghvl{{(\ell-1)}}+\mathbf{b}_{2}^{(\ell)},\\
&\gdvinl= \sum_{\forall u\in N_{in}(v)}\mathbf{W}_{3}^{(\ell)}(\gzul{{(\ell)}}||(\gzvl{{(\ell)}}-\gzul{{(\ell)}})),\\
&\gdvoutl = \sum_{\forall u\in N_{out}(v)}\mathbf{W}_{3}^{(\ell)}(\gzul{{(\ell)}}||(\gzvl{{(\ell)}}-\gzul{{(\ell)}})),\\
&\ghvl{{(\ell)}} =  \awv{1}\gzvl{{(\ell)}}+\awv{2}\gdvinl + \awv{3}\gdvoutl,\\
\end{split}
\end{equation}
where $\Nin{v}$ and $\Nout{v}$ are the multisets of $v$'s incoming and outgoing neighbors respectively; $\mathbf{W}_{2}^{(\ell)} \in\mathbb{R}^{c\times c}$, $\mathbf{b}_{2}^{(\ell)}\in\mathbb{R}^{c}$, and  $\mathbf{W}_{3}^{(\ell)}\in\mathbb{R}^{c\times 2c}$ are learnable parameters; $\awv{1}$, $\awv{2}$, and $\awv{3}$ are attention weights.

Attentions $\awv{1}$, $\awv{2}$, and $\awv{3}$ are calculated by Eq. \eqref{eq:attention}. A larger attention weight indicates that the corresponding aspect is more important in the message passing process, which provides a flexible way to aggregate the messages in Eq. \eqref{eq:gdl}. 

\begin{equation} \label{eq:attention}
\begin{split}
& {w}_{v,1} = \sigma(\gzvl{{(\ell)}}\cdot\mathbf{q}); {w}_{v,2} = \sigma(\gdvinl\cdot\mathbf{q}); {w}_{v,3} = \sigma(\gdvoutl\cdot\mathbf{q}), \\
& \awv{k}=\mathrm{softmax}(({w}_{v,1},{w}_{v,2},{w}_{v,3}))_k,
\end{split}
\end{equation}
where $\sigma$ is LeakyReLU activation, $\mathbf{q} \in\mathbb{R}^{c}$ is the learnable attention vector, softmax is a normalization, and $k=1,2,3$.

\header
\textbf{Discussion.}
There are several ways to handle the discrepancy issue in literature.
Here we highlight the technical differences of MGD compared with  existing work \cite{zhang2021fraudre,ding2021inductive,dou2020enhancing,wang2021decoupling,liu2020alleviating}.
Moreover, we experimentally compare MGD with these methods in Section \ref{sec:experiments}.
Compared with our MGD, the counterpart in \fraudre, dubbed as FRA, (Eq. (2) in \cite{zhang2021fraudre}) does {not} have the latter two designs in MGD and uses mean pooling. As analyzed in \cite{xu2018powerful}, sum pooling yields higher expressive power than mean pooling, particularly for {multiset} neighborhoods of multigraphs in this paper. 
Further, the attention mechanism and learnable layer-wise transformations in MGD enable the
flexible pass and aggregation of both incoming and outgoing discrepancy-aware messages along parallel edges. 
Thus, MGD is technically different from  \fraudre.
In \cite{ding2021inductive}, GDN \textit{only} aggregates the representation differences between a target node and its neighbors, while {omitting} neighbor representations themselves (Eq. (1) and (2) in \cite{ding2021inductive}). 
Contrarily, our MGD passes richer messages containing \textit{both} neighbor discrepancies and neighbor representations.
There are also different methodologies in \cite{dou2020enhancing,liu2020alleviating, wang2021decoupling}.
In \cite{dou2020enhancing,liu2020alleviating}, they train samplers to identify discrepant neighbors, \eg via reinforcement learning in \cite{dou2020enhancing}. 
DCI \cite{wang2021decoupling} adopts self-supervised learning and clustering to decouple representation learning and classification.
In experiments, \algo outperforms these existing methods for illicit account detection on directed multigraphs with edge attributes, validating the effectiveness of our designs in MGD.

%% file: experiments.tex
\section{Experiments}
\label{sec:experiments}
We experimentally evaluate   \algo against 15 baselines on 4 real-world transaction networks of  cryptocurrency datasets, with the aim to answer the following 5 research questions: 
\begin{itemize}
    \item \textbf{RQ1:} How does \algo perform in terms of effectiveness, compared with existing state of the art?
    \item \textbf{RQ2:} How does the MGD module perform, compared with existing counterparts?
    \item \textbf{RQ3:} How does the \edgeseq module perform, compared with manual feature engineering?
    \item \textbf{RQ4:} How is the training efficiency of \algo?
    \item \textbf{RQ5:} How does \algo perform in sensitivity analysis?
\end{itemize}

\subsection{Experimental Setup} \label{sec:expsetup}

\noindent
\textbf{Datasets.}
We evaluate on 4 large cryptocurrency datasets, including 2 Ethereum datasets and 2 Bitcoin datasets. The statistics of the datasets are listed in Table \ref{tab:datasets}. The first three datasets are from existing works, and we create the last largest Bitcoin dataset with more than 20 million nodes and 203 million edges.  {We obtain ground-truth labels  of the datasets by crawling   illicit and normal account labels from reliable sources, including Etherscan~\cite{etherscan_2022} and WalletExplorer~\cite{walletexplorer_2022}.}   
{\ethsecond} \cite{yuan2020phishing} and {\ethphish} \cite{chen2020toit} are two Ethereum transaction networks. In both datasets, every edge has two attributes: transaction amount and timestamp.
The labeled illicit nodes are the addresses that conduct phishing scams in these two datasets. 
For \ethphish dataset from \cite{chen2020toit}, it only contains illicit node labels. We enhance the dataset by identifying the benign accounts (\eg wallets and finance services) in \ethphish from Etherscan \cite{etherscan_2022}  as normal node labels.
{\bitcoinS} \cite{wu2021detecting} contains the first 1.5 million transactions from June 2015. 
As explained in Section \ref{sec:preliminary}, a Bitcoin transaction can involve multiple senders and receivers.
After built as a multigraph, \bitcoinS has about 2.5 million nodes and 14 million edges.
In \bitcoinS, an edge has 5 attributes: input amount, output amount, number of inputs, number of outputs, and  timestamp.
We build the largest {\bitcoinL} based on all transactions happened from June to September 2015. \bitcoinL has more than 20 million nodes and 200 million edges, and each edge has 8 attributes: input amount, output amount, number of inputs, number of outputs, fee, total value of all inputs, total value of all outputs, and timestamp. 
We obtain the labeled data in \bitcoinS and \bitcoinL   by crawling from WalletExplorer~\cite{walletexplorer_2022}.
Following \cite{wu2021detecting}, Bitcoin addresses belonging to gambling and mixing services are regarded as illicit accounts due to their strong association with money laundering activities, while the addresses in other types are normal accounts.
Parallel edges between nodes are common in the datasets. For instance, in \ethphish, there are 5,353,834 connected node pairs, and 1,287,910 of them have more than one edge (24.06\%).

\begin{table}[!t]
\centering  
  \caption{Statistics of the datasets.}
  \label{tab:datasets}
  \vspace{-3mm}
  \small
  \setlength{\tabcolsep}{1pt}
  \resizebox{1\linewidth}{!}{ 
  \begin{tabular}{lllclll}
    \toprule
    Dataset&\#Nodes&\#Edges&\#Edge attribute&\#Illicit&\#Normal&Illicit:Normal\\
    \midrule
    \ethsecond \cite{yuan2020phishing} & 1,329,729 & 6,794,521 & 2 & 1,660 & 1,700 & 1:1.02\\
    \ethphish \cite{chen2020toit} & 2,973,489 & 13,551,303 & 2 & 1,165 & 3,418 & 1:2.93\\
    \bitcoinS \cite{wu2021detecting} & 2,505,841 & 14,181,316 & 5 & 46,930 & 213,026 & 1:4.54\\
    \bitcoinL & 20,085,231 & 203,419,765 & 8 & 362,391 &1,271,556 & 1: 3.51\\
  \bottomrule
\end{tabular}
}
\vspace{-3mm}
\end{table}

\begin{table*}[tp] 
\addtolength{\tabcolsep}{-1pt}
\renewcommand{\arraystretch}{0.95}
\caption{Overall results on all datasets (in percentage \%). \textbf{Bold}: best. \underline{Underline}: runner-up. Relative improvements by \algo over runner-ups in brackets.}  
\label{tab:overall} 
\vspace{-1mm}

\resizebox{0.92\textwidth}{!}{ 
\setlength{\tabcolsep}{1pt} 
  \centering  
  \begin{threeparttable}
    \resizebox{\linewidth}{!}{ 
    \begin{tabular}{l|cccc|cccc|cccc|cccc}  
    \toprule  
    \multirow{2}{*}{Method}&  
    \multicolumn{4}{c|}{\ethsecond}&\multicolumn{4}{c|}{\ethphish}&\multicolumn{4}{c|}{\bitcoinS}&\multicolumn{4}{c}{\bitcoinL}\cr
    \cmidrule(lr){2-5} \cmidrule(lr){6-9} \cmidrule(lr){10-13} \cmidrule(lr){14-17}
    &Precision&Recall&F1&AUC&Precision&Recall&F1&AUC&Precision&Recall&F1&AUC&Precision&Recall&F1&AUC\cr  
    \midrule  
    \textbf{\gcn}  & 81.21  & \underline{96.35}  & 88.09  & 87.52 & 86.07  & 80.15  & 82.97  & 87.98  & 79.90  & 81.21  & 80.49  & 88.33  & 80.11  & 83.35  & 81.68  & 88.72  \\
    \textbf{Sage} & \underline{92.92} & 89.95  & \underline{91.39} & \underline{91.71} & \underline{90.49} & 91.14  & \underline{90.81 } & 94.04 & \underline{87.17} & \underline{83.27} & \underline{85.16} & \underline{90.28} & 83.16  & \underline{84.79 } & \underline{83.92} & \underline{89.93} \\
    \textbf{GAT}   & 85.99  & 93.55  & 89.60  & 89.52  & 85.11  & 85.37  & 85.06  & 90.23 & 86.16  & 81.45  & 83.71  & 89.27  & 79.45  & 65.73  & 71.80  & 80.44  \\
    \textbf{GATE} & 66.49  & 90.66  & 76.70  & 73.62  & 88.66  & 85.41  & 86.98  & 90.95 & 71.28  & 67.06  & 68.96  & 80.52  & 67.76  & 36.86  & 47.58  & 65.87  \\
    \textbf{GINE} & 75.65  & 88.63  & 81.58  & 80.66  & 82.45  & 81.75  & 82.02  & 88.07  & 64.68  & 61.88  & 63.14  & 77.17  & 70.06  & 55.45  & 61.70  & 74.27  \\
    \textbf{TransConv} & 90.97  & 86.16  & 88.47  & 89.00  & 84.92  & 91.25  & 87.95  & 93.03  & 70.55  & 56.43  & 62.62  & 75.61  & 73.93  & 64.09  & 68.52  & 78.79  \\
    \midrule  
    \textbf{GDN}   & 85.55  & 83.79  & 84.63  & 85.14  & 82.42  & 84.00  & 83.19  & 89.13 & 81.56  & 74.76  & 77.99  & 85.51  & {73.68}  & {45.92}  & {56.57}  & {70.62} \\
    \textbf{CARE-GNN}  & 79.81  & 88.53  & 83.93  & 83.61  & 72.72  & 82.76  & 77.41  & 86.42  & 33.41  & 71.93  & 45.36  & 70.04  & 31.02  & 73.29  & 43.57  & 63.50  \\
    \textbf{DCI}   & 71.77  & 92.18  & 80.71  & 78.86  & 75.44  & 76.95  & 76.19  & 84.47 & 82.19  & 51.47  & 63.30  & 74.50  & 81.18  & 51.77  & 62.53  & 73.91  \\
    \textbf{PC-GNN}  & 83.34 & 81.28 & 82.26 & 82.87 & 79.18 & 89.38 & 83.96 & 90.93 & 36.89 & 73.01 & 48.87 & 72.72 & 30.04 & 82.68 & 44.07& 64.01 \\
    \textbf{FRAUDRE}
     & 73.3  & 95.26 & 82.83 & 81.1  & 84.06 & 80.28 & 82.10 & 82.89 & 36.05 & 72.97 & 48.23 & 72.27 & 
    33.46 & 76.67 & 46.59 & 66.69 \\
    \midrule  
    \textbf{SigTran} & 87.10  & 93.33  & 90.11  & 90.23  & 69.41  & 55.47  & 61.66  & 73.90  & 75.97  & 52.24  & 61.91  & 74.30  & 83.25 & 75.22  & 79.03  & 85.45  \\
    \textbf{Pdetector} & 79.27  & 91.60  & 84.99  & 84.65  & 82.37  & 83.58  & 82.97  & 88.98  & 80.43  & 58.77  & 67.92  & 77.81  & 77.08  & 52.76  & 62.64  & 74.14  \\
    \textbf{EdgeProp}  & 81.59  & 85.36  & 83.30  & 83.38  & 89.49  & \underline{91.78 } & 90.57  & \underline{94.15 } & 73.82  & 69.21  & 71.39  & 81.88  & 72.39  & 67.09  & 69.51  & 79.84  \\
    \textbf{BERT4ETH}  &  85.54 & 87.65  & 86.58  & 86.93  & 88.35  & 80.29 & 84.13  & 88.48 & 80.03  & 59.95  & 68.55  & 78.32  & \underline{84.70}  &  77.36 & 80.87  & 86.69  \\    
    
    \midrule 
    \multirow{2}{*}{\textbf{\algo}} & \textbf{97.11 } & \textbf{96.68 } & \textbf{96.89 } & \textbf{96.97 } &
    \textbf{94.82} & \textbf{92.95} & \textbf{93.86} & \textbf{95.66}& \textbf{92.83 } & \textbf{90.39 } & \textbf{91.59 } & \textbf{94.43 } & \textbf{97.72 } & \textbf{95.40 } & \textbf{96.55 } & \textbf{97.39 } \\
    & (+4.5\%) & (+0.3\%) & (+6.0\%) & (+5.7\%) &(+4.8\%) & (+1.3\%) & (+3.4\%) & (+1.6\%)& (+6.5\%) & (+8.6\%) & (+7.6\%) & (+4.6\%) & (+15.4\%) & (+12.5\%) & (+15.1\%) & (+8.3\%) \\

    \bottomrule  
    \end{tabular}  
}
 
 \end{threeparttable}  

 }
 \vspace{0mm}
\end{table*}  

\header
\textbf{Baselines.}
We compare with  15 competitors in  3 categories, which are reviewed in Section \ref{sec:relatedwork}.

\begin{itemize}[leftmargin=*]
    \item \textit{Cryptocurrency illicit account detection methods,} including {\pdetector} \cite{chen2020toit}, {\sigtran} \cite{poursafaei2021sigtran}, {\edgeprop} \cite{tam2019identifying}, {\bert} \cite{hu2023bert4eth}.

\item \textit{Graph-based anomaly detection methods,} including  {\caregnn} \cite{dou2020enhancing}, {\dci} \cite{wang2021decoupling}, {\pc}\cite{liu2021pick}, {\gdn} from AEGIS \cite{ding2021inductive}, and {\fraudre} \cite{zhang2021fraudre}. Specifically, the baseline \gdn is a message passing module in AEGIS, while AEGIS itself is unsupervised and thus not compared in the supervised setting.

\caregnn, \pc, and \fraudre are designed for relation graphs, and we set the number of relations as 1, to run them.
\item \textit{GNN models,} including {\gcn} \cite{kipf2017semi}, {\sage} \cite{hamilton2017inductive}, {\gat} \cite{velivckovic2017graph}, {\gate} \cite{velivckovic2017graph}, {\gine} \cite{hu2019strategies}, and {\transconv} \cite{shi2020masked}.

\end{itemize}

\header
\textbf{Implementation Details.}
We implement \algo and GNN-based models using Pytorch  and Pytorch Geometric. We also use Pytorch to implement \gdn and \pdetector following the respective papers. For the other competitors, we use the codes provided by the authors. 
All experiments are conducted on a Linux server with Intel Xeon Gold 6226R 2.90GHz CPU and an Nvidia RTX 3090 GPU card. 
For the baselines requiring initial node features as input, following the way in \cite{poursafaei2021sigtran}, we obtain node features, such as node degree and total received/sent amount, by feature engineering for the baselines. Particularly, in this way, we get 48, 48, 69, and 89 node features for datasets \ethphish, \ethsecond, \bitcoinS, and \bitcoinL respectively. In terms of \pdetector, we extract the 8 specific node features suggested in its paper \cite{chen2020toit} for its training to make a fair comparison.
\gdn, \edgeprop, as well as the GNN-based models, are not originally designed for the binary classification task in this paper. Therefore, we regard them as the encoder to generate node representations, which are then sent to a 2-layer MLP classifier with the same objective in Section \ref{sec:obj}.

\header
\textbf{Parameter Settings.}
We set node representation dimension ($c=128$), the number of GNN layers (2), learning rate (0.001), dropout rate (0.2).
In \algo, we set maximum sequence length $T_{max}=32$.
We will study the impact of $T_{max}$ in Section~\ref{sec:ablation}.
For all methods, we adopt Adam optimizer, mini-batch training~\cite{hamilton2017inductive} with batch size 128, and, if not specified, rectified linear units (ReLU) is used as the activation function. 
For all GNN models, \gdn, \edgeprop, and our method that require neighborhood sampling, given a target node, we randomly sample its 1 and 2-hop neighbors with sample size 25 and 10 respectively. 
For other settings in baselines, we follow the instructions in their respective papers. The number of training epochs is set as 30 in \ethsecond, \ethphish,  and \bitcoinS, and set as 10 in \bitcoinL.

\header
\textbf{Evaluation Settings.}
We adopt 4 evaluation metrics: Precision, Recall, F1 score, and Area Under ROC curve (AUC for short).
All metrics indicate better performance when they are higher.
For each dataset, we split all labeled nodes  into training, validation, and testing sets with ratio 2:1:1.
Each model is trained on the training set. When a model achieves the highest F1 score on the validation set, we report the evaluation results on the testing set as the model's performance. 
For each method, we train it for 5 times and report the average value of each evaluation metric. We also study  the training time and the impact when varying training set size as well as the percentage of illicit node labels.

\subsection{Overall Effectiveness} \label{sec:overalleffectiveness}
To answer RQ1, we report the overall results of   \algo and all competitors on all datasets in
Table \ref{tab:overall}.
First, observe that \algo consistently achieves the highest accuracy by all evaluation metrics over all datasets, outperforming all baselines often by a significant margin.  
For instance, on \ethsecond, \algo achieves $96.89\%$ F1 score, while the F1 of the best competitor \sage is $91.39\%$, indicating a relative improvement of $6\%$.
On \ethphish, \algo has precision $94.82\%$, outperforming the best competitor by a relative improvement of $4.8\%$.
On \bitcoinS and \bitcoinL, \algo also achieves the highest accuracy for illicit account detection. In particular, \algo achieves 91.59\% and 96.55\% F1 scores on \bitcoinS and \bitcoinL, 7.6\% and 15.1\% relatively higher than the best baselines, respectively.  
Another observation is that the performance gain of \algo is larger on the largest  \bitcoinL, \eg 15.4\% precision improvement over the best competitor \sigtran as shown in Table \ref{tab:overall}.  The reason is that \algo with \edgeseq is able to take advantage of the abundant edge attributes in the multigraph of  \bitcoinL,  to automatically extract  informative representations for accurate detection of illicit accounts.
Existing solutions, such as \sigtran, require manual feature engineering, and thus, could not effectively leverage the  large-scale data to preserve the intrinsic transaction patterns of accounts. In Section \ref{sec:expGdl}, we conduct  an evaluation to further reveal the effectiveness of \edgeseq, compared with handcrafted features.
We conclude that  \algo achieves superior performance for illicit account detection on cryptocurrencies.

\subsection{Study on MGD and Edge2Seg} \label{sec:expGdl}

\begin{figure}[t!]
	\centering
		\includegraphics[width=0.95\columnwidth]{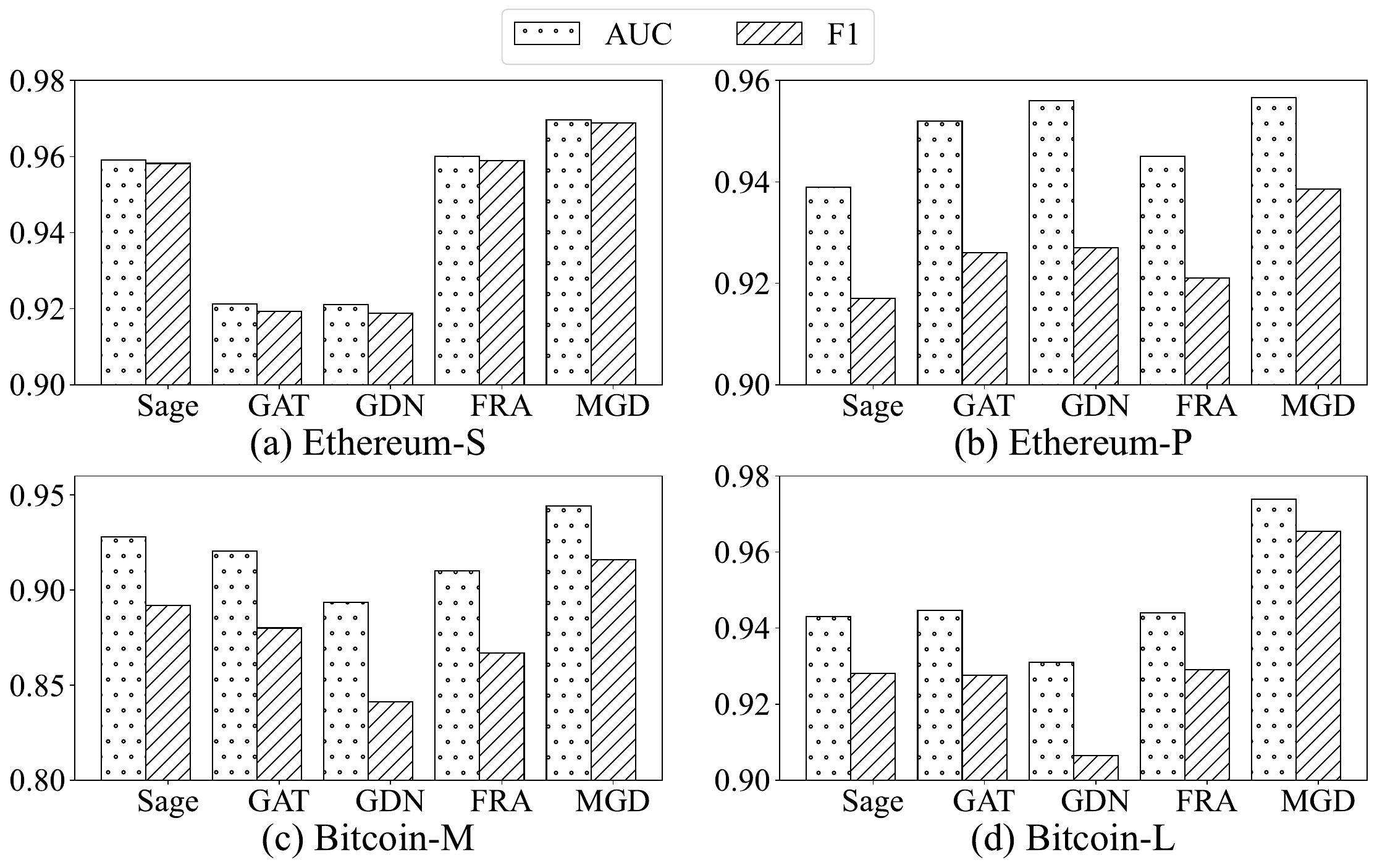}
\vspace{-2mm}
  
\caption{{Compare the  {MGD} module  with other GNN layers.} }
\vspace{-2mm}
\label{fig:ablation}
\end{figure}

\header
\textbf{MGD Evaluation.}
As we have discussed in Section \ref{sec:gdl}, our  MGD is different from existing work. To test the effectiveness of MGD in \algo and answer RQ2,  we replace MGD with existing GNN layers, namely, Sage layer~\cite{hamilton2017inductive}, GAT layer~\cite{vaswani2017attention}, GDN layer in AEGIS~\cite{ding2021inductive}, and FRA layer in \fraudre~\cite{zhang2021fraudre}, and compare their performance.
Figure \ref{fig:ablation} presents the F1 and AUC results for \algo across all datasets, using each of the five different GNN layers.
Observe that \algo with MGD always achieves the highest F1 and AUC scores on all datasets, and outperforms \gdn, \sage, \gat, and FRA layers.
The results demonstrate the effectiveness of our MGD to preserve the differentiable representations of both illicit and benign nodes with the consideration of the discrepancies when conducting message passing over the multigraph topology.
In particular, given a target  node $v$, \sage and \gat layers do not consider discrepancies, \gdn layer only passes and aggregates the representation differences of its neighbors to it.
Compared with the FRA layer, our MGD employs sum pooling, layer-wise learnable transformations, and an attention mechanism to flexibly pass and aggregate both incoming and outgoing neighbor discrepancies and neighbor representations.
Moreover, among existing GNN layers,  \gdn layer performs better than \sage, \gat, and FRA layers on \ethphish in Figure \ref{fig:ablation}(b), while being inferior on  the other three datasets.
This indicates that it is also important to propagate and aggregate neighbor representations to target nodes in the input multigraph, rather than only considering node representation differences, for effective illicit account detection.

\begin{table}[t!]
\renewcommand{\arraystretch}{0.85}
  \centering
    \caption{
  \textbf{Manual} features v.s., learned representations by \textbf{\edgeseq} on  \bitcoinL (in percentage \%). Relative improvements of \edgeseq over Manual are in brackets.}
  \vspace{-2mm}

  \small
  \resizebox{0.85\columnwidth}{!}{ 
    \begin{tabular}{clll}
    \toprule
    \multicolumn{1}{l}{GNN Layer} & \multicolumn{1}{l}{Variant} & \multicolumn{1}{l}{F1} & \multicolumn{1}{l}{AUC} \\
    \midrule
    \multirow{2}[4]{*}{Sage} & {Manual} & 83.92  & 89.93  \\
\cmidrule{2-4}          & {\edgeseq} & 92.80 (+10.6\%) & 94.30 (+4.9\%) \\
    \midrule
    \multirow{2}[4]{*}{\gat} & {Manual} & 71.80  & 80.44 \\
\cmidrule{2-4}          & {\edgeseq} & 92.75 (+29.2\%)  & 94.46 (+17.4\%)  \\

    \midrule
    \multirow{2}[4]{*}{MGD} & {Manual} & 85.29  & 92.39 \\
\cmidrule{2-4}          &
{\edgeseq} & 96.55 (+13.2\%)  & 97.39 (+5.4\%)  \\
    \bottomrule
    \end{tabular}
    }
  \label{tab:addlabel}
  \vspace{0mm}
\end{table}

\header
\textbf{Edge2Seq Evaluation.}
To answer RQ3, we demonstrate the power of \edgeseq by interchanging it with the handcrafted features as the input of \sage, \gat, and our MGD, and report the evaluation results on \bitcoinL in Table \ref{tab:addlabel}.
Specifically, in Table \ref{tab:addlabel}, {Manual} indicates to have the handcrafted node features introduced in Section~\ref{sec:expsetup} as the initial input of node representations for training, while {\edgeseq} automatically learns node representations by applying GRUs over the incoming and outgoing edge sequences of nodes.
As shown in Table \ref{tab:addlabel}, comparing against \sage (resp. \gat) with manual features, \sage (resp. \gat) with \edgeseq always achieves higher F1 and AUC by a significant margin. For instance, \gat with \edgeseq improves \gat with manual features by a significant margin of 29.2\%.
The results indicate the superiority of \edgeseq, compared with manual feature engineering.
Further, the result of our MGD with manual features in Table \ref{tab:addlabel} (\ie \algo without \edgeseq) also indicates that \edgeseq is important for the problem studied in this paper.
Our method \algo assembling \edgeseq and MGD together obtains the best performance, as shown in Table \ref{tab:addlabel}.

\begin{table}[!t]
\renewcommand{\arraystretch}{0.85}
\vspace{0mm}
    \renewcommand\tabcolsep{2pt}
  \centering
  \caption{Training time per epoch (Seconds)}
  \vspace{-2mm}
  \small
  \resizebox{0.8\columnwidth}{!}{ 
    \begin{tabular}{l|cccc}
    \toprule
    Method & \multicolumn{1}{l}{\ethsecond} & \multicolumn{1}{l}{\ethphish} & \multicolumn{1}{l}{\bitcoinS} & \multicolumn{1}{l}{\bitcoinL} \\
    \midrule
    \gcn   & 0.29  & 0.65  & 14.47  & 274.20  \\
    \sage  & 0.31  & 0.67  & 13.76  & 270.80  \\
    \gat   & 0.75  & 1.14  & 18.25  & 307.70  \\
    \gate  & 0.57  & 0.92  & 13.42  & 130.08  \\
    \gine  & 0.13  & 0.39  & 8.92  & 104.23  \\
    \transconv & 0.22  & 0.66  & 14.72  & 181.82  \\
    \midrule
    EdgeProp & 0.17  & 0.39  & 9.73  & 105.22  \\

    GDN   & 0.39  & 0.74  & 18.47  & 294.64  \\
    CARE-GNN & 0.62  & 1.80  & 29.19  & 257.45  \\
    DCI   & 1.06  & 1.37  & 46.27  & 972.52  \\
    PC-GNN & 1.62  & 6.10  & 90.70  & 6525.68  \\
    FRAUDRE & 1.34  & 2.58  & 75.40  & 884.20  \\
    \bert & 2.57 & 3.69 & 286.32 & 4107.06 \\
    \textbf{\algo} & {0.45} & {0.70} & {35.92} & {330.73} \\
    \bottomrule
    \end{tabular}
  }
  \label{tab:efficiency}
  \vspace{0mm}
\end{table}

\subsection{Training Efficiency}
\label{sec:efficiency}
To answer RQ4, Table \ref{tab:efficiency} reports the average training time per epoch of \algo and the competitors in seconds on all datasets.
First, observe that  anomaly detection methods (\gdn, \caregnn, \dci, \pc, and \fraudre) and our method \algo are generally slower than the common GNN models listed in the first group of Table \ref{tab:efficiency}, \eg \gcn and \sage, which is because of the unique designs for illicit/anomaly detection in these methods. However, as reported in Section \ref{sec:overalleffectiveness},  compared with \algo, common GNN models yield inferior accuracy since they
are not dedicated to the task of illicit account detection.
Second, \algo is faster than most graph-based anomaly detection methods. Specifically, on \ethsecond and \ethphish, \algo is faster than \caregnn, \dci, \pc, and \fraudre.
On \bitcoinS and \bitcoinL, \algo is faster than \dci, \pc, and \fraudre. In addition, although \edgeprop is fast, it is not as accurate as \algo as shown in Section \ref{sec:overalleffectiveness}.
The training time per epoch in Table \ref{tab:efficiency} does not include  \sigtran and \pdetector, since they are not trained in an epoch manner.
Considering together the training efficiency in Table \ref{tab:efficiency} and the effectiveness in Table \ref{tab:overall}, we can conclude that \algo has superior accuracy for illicit account detection, while being reasonably efficient, on large-scale 
 cryptocurrency datasets.

\subsection{Sensitivity Analysis}
\label{sec:ablation}
We conduct  experiments for sensitivity analysis to answer RQ5.

\begin{figure}[t!]
\centering    
\subfigure[F1-score]
{
\includegraphics[width=0.42\linewidth]{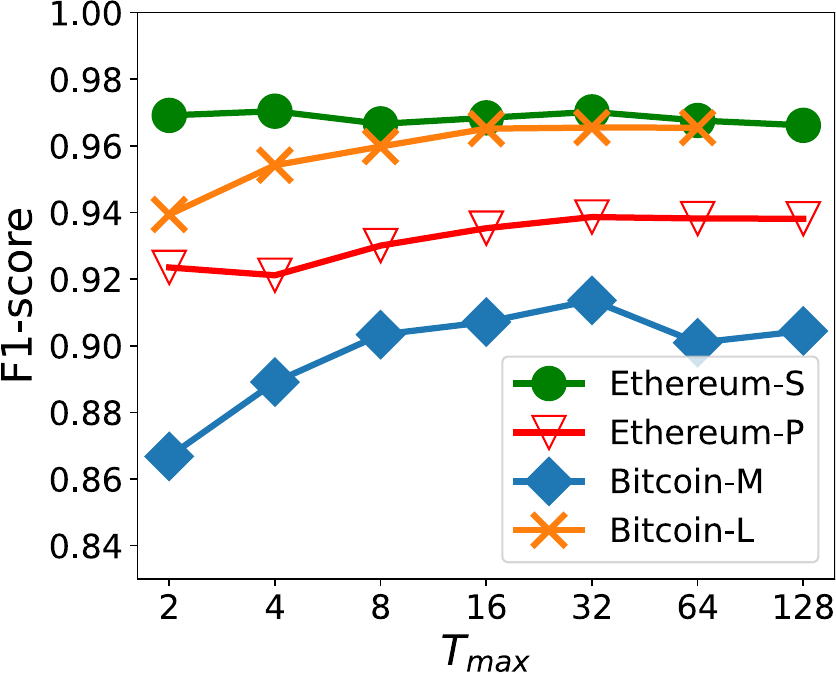} 
\label{fig::length-f1}
}
\subfigure[Training time]
{
\includegraphics[width=0.42\linewidth]{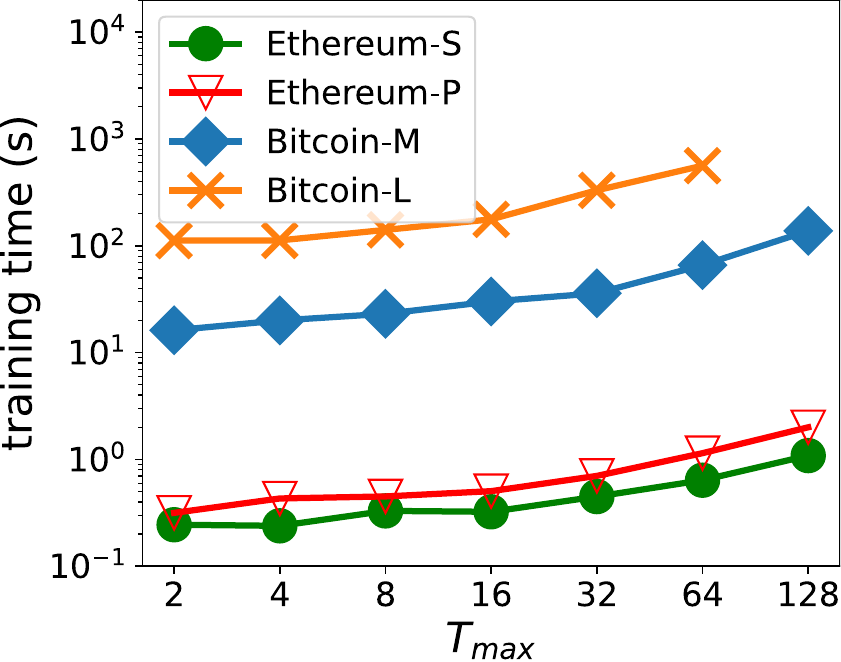} 
\label{fig::length-time}
}
\vspace{-3mm}
\caption{Performance Comparison  with Varying $T_{max}$}
\label{fig:length}
\vspace{-2mm}
\end{figure}

\header
\textbf{Varying the maximum sequence length $T_{max}$.}
We vary $T_{max}$ in \edgeseq from 2 to 128 and report the performance of \algo and average training time per epoch (seconds) in Figure \ref{fig:length}. 
The result of $T_{max}=128$ on \bitcoinL is not reported due to out of GPU memory.
In Figure \ref{fig::length-f1}, observe that as $T_{max}$ increases, F1 score on \ethsecond is relatively stable, F1 score on \ethphish and 
\bitcoinL increases first and then becomes stable, and F1 score on \bitcoinS increases first and then decreases after $T_{max}$ is beyond 32. As discussed in \cite{liu2021intention}, the decrease in \bitcoinS may be caused by the noise introduced among distant elements when considering very long sequences in sequence models. 
Therefore, we choose $T_{max}=32$ as default in experiments.
In terms of training time per epoch in Figure \ref{fig::length-time}, when $T_{max}$ increases, it takes more time for training on all datasets, which is intuitive since there are longer sequences to be handled by \edgeseq. The increasing trend of training time is consistent with the time complexity analysis   in Section \ref{sec:obj}.

\begin{table}[!t]  
\addtolength{\tabcolsep}{-2pt}
\caption{Ablation Study (in percentage \%)}  
\label{tab:ablation}  
\vspace{-1mm}
  \centering  
  \begin{threeparttable}  
    \footnotesize
    \begin{tabular}{c|cc|cc|cc|cc}  
    \toprule  
    \multirow{2}{*}{Methods}&  
    \multicolumn{2}{c|}{\ethsecond}&\multicolumn{2}{c|}{\ethphish}&\multicolumn{2}{c|}{\bitcoinS}&\multicolumn{2}{c}{\bitcoinL}\cr
    \cmidrule(lr){2-3} \cmidrule(lr){4-5} \cmidrule(lr){6-7} \cmidrule(lr){8-9}
    &F1&AUC&F1&AUC&F1&AUC&F1&AUC\cr  
    \midrule  
    \textbf{\algo\textbackslash{}MGD}  & 95.17  & 95.26   & 82.23 & 88.66 & 69.81  & 81.87  & 72.96  & 81.77  \\
    \textbf{\algo\textbackslash{}A} & 96.75  & 96.83  & 93.28  & 95.62  & 89.94  & 92.93  & 95.36  & 96.65  \\
\midrule  
    \textbf{\algo} & \textbf{97.11 } & \textbf{96.97 } & \textbf{93.86} & \textbf{95.66} & \textbf{91.59 } & \textbf{94.43 } & \textbf{97.72 } & \textbf{97.39 } \\
    
    \bottomrule  
    \end{tabular}  
 \end{threeparttable}
 \vspace{-2mm}
\end{table}

\header 
\textbf{Ablation Study.} To validate the effectiveness of every component in \algo, we conduct extra ablation study by evaluating \algo without MGD in Section \ref{sec:gdl} (denoted as \algo\textbackslash{}MGD), and \algo without the attention mechanism in Eq. \eqref{eq:attention} (\ie set $\awv{1}=\awv{2}=\awv{3}=1$ in Eq. \eqref{eq:gdl}), denoted as \algo\textbackslash{}A.
Table \ref{tab:ablation} presents their performance compared with the complete version \algo. 
First, observe that the performance on all four datasets increases as we add more techniques, validating the effectiveness of the proposed MGD and attention mechanism.
Further, note that essentially \algo\textbackslash{}MGD is only with \edgeseq (\ie only considering a node's local transaction features), and thus, it has inferior performance   as shown in Table \ref{tab:ablation}. This observation indicates the importance of incorporating the multigraph topology for illicit account detection.

\begin{figure}[t!]
	\centering
		\includegraphics[width=0.92\linewidth]{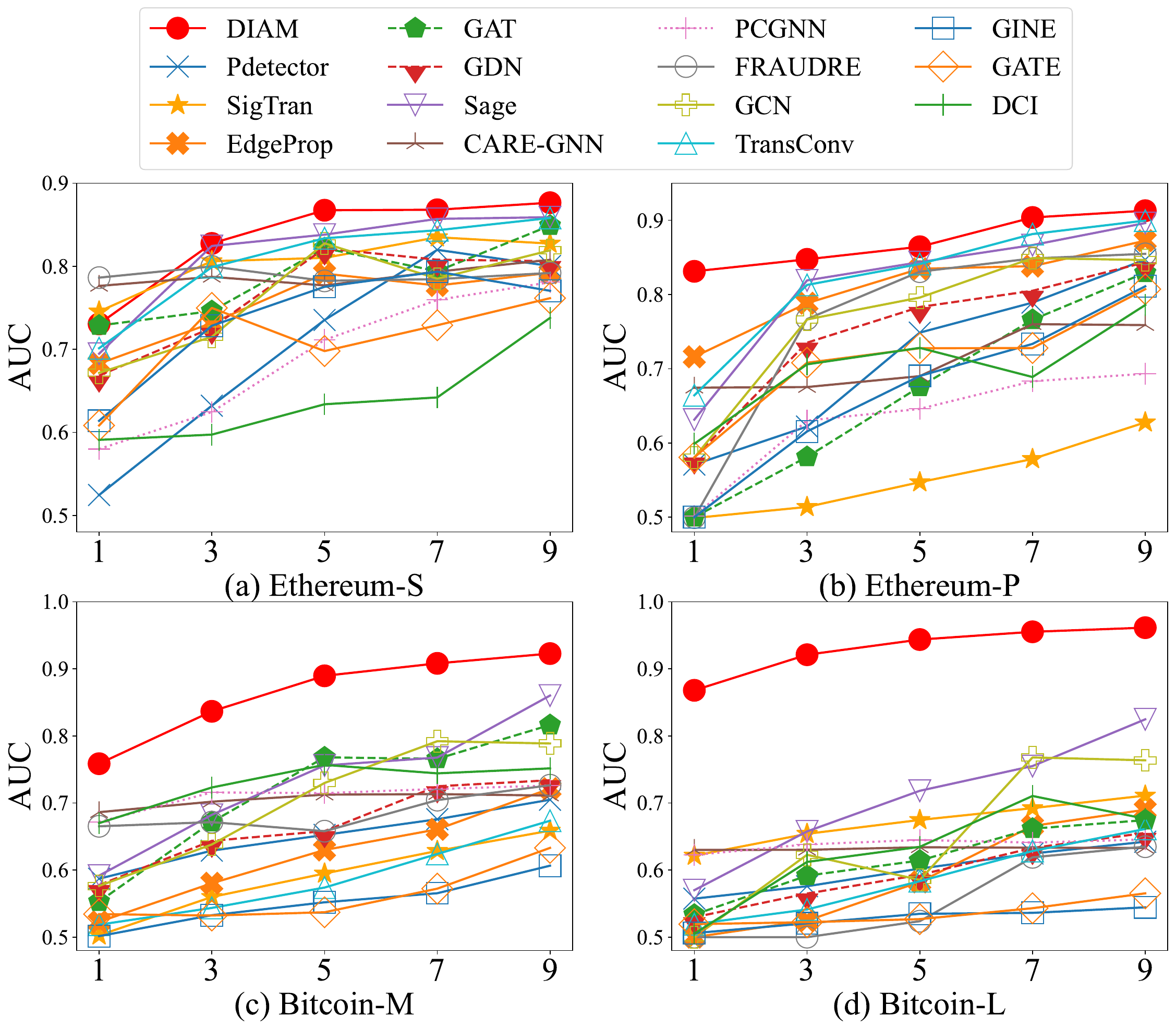}
  \vspace{-2mm}
\caption{Varying {illicit} ratio (\%) on all datasets.
}
\label{fig:anomaly_rate}   
\vspace{-3mm}
\end{figure}

\header 
\textbf{Varying illicit ratio.} 
As shown in Table \ref{tab:datasets}, the number of illicit accounts is relatively high compared with normal nodes, particularly on \ethsecond and \ethphish  datasets. 
In order to stress test \algo and the baselines when the illicit node labels are scarce, we have conducted  experiments to vary the illicit ratio from 1\% to 9\%, by randomly sampling a subset of illicit nodes in training on every dataset.
The illicit ratio is the proportion of illicit nodes in all labeled training nodes. 
Figure \ref{fig:anomaly_rate} reports the performance of all methods on all datasets.
The overall observation is that \algo outperforms existing methods under most illicit ratios, except the AUC at 1\% on \ethsecond. 
As the illicit ratio decreases, the performance of all methods drops on all datasets, since all methods would be under-trained with limited labels. 
Further, the superiority of \algo is more obvious on larger datasets. The reason is that our method can better leverage the abundant data to automatically extract meaningful features via \edgeseq and MGD in \algo.
The  results in Figure \ref{fig:anomaly_rate} demonstrate the effectiveness of the proposed \algo when labels are scarce.

\begin{figure}[t!]  
\centering    
\subfigure[\ethsecond]
{
	\label{fig::rate_EthereumS}
	\includegraphics[width=0.4\linewidth]{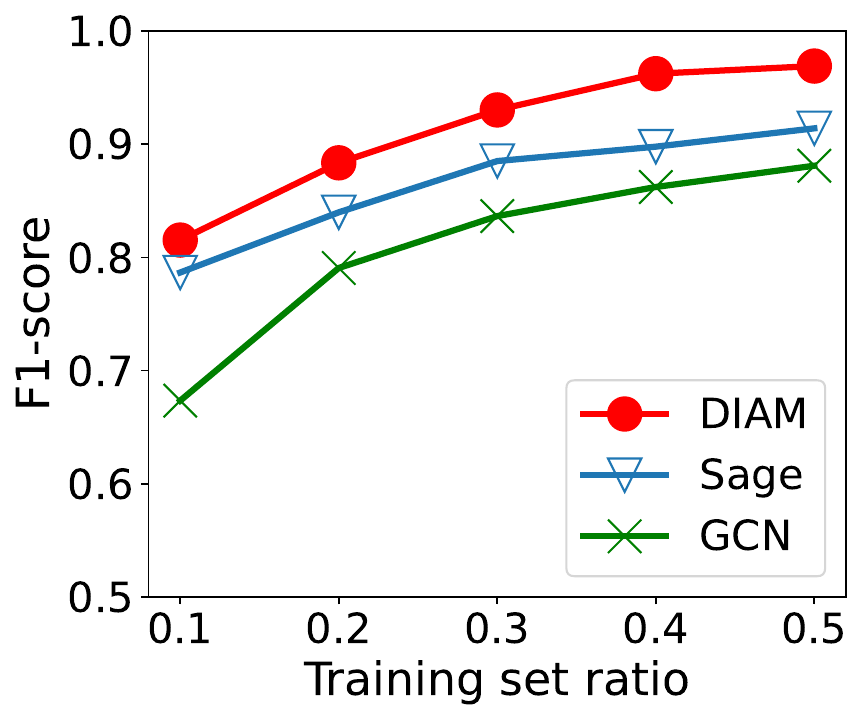} 
}
\subfigure[\ethphish]
{
	\label{fig::rate_EthereumP}
	\includegraphics[width=0.4\linewidth]{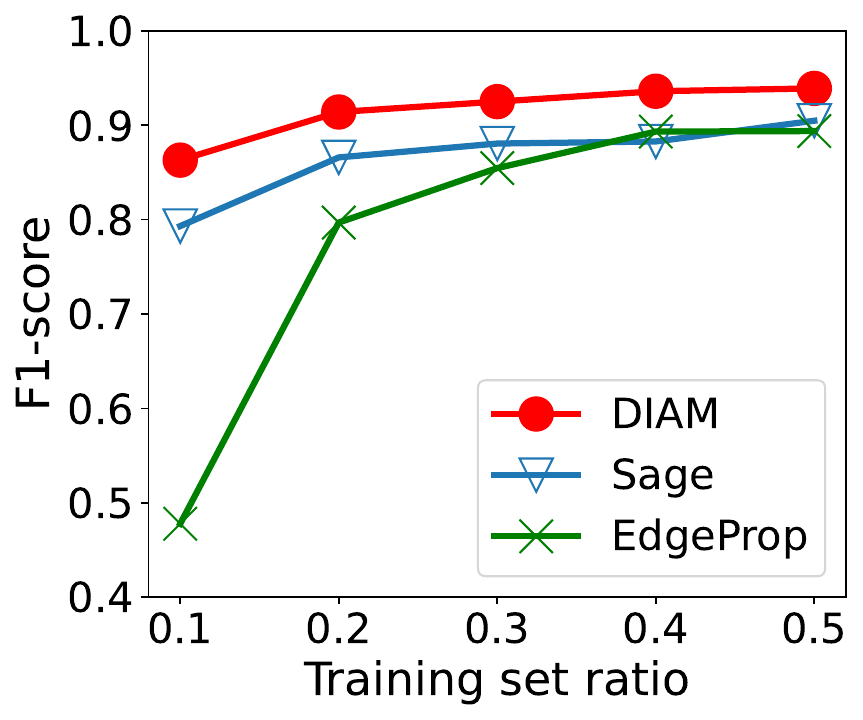} 
}
\vspace{-2pt}
\subfigure[\bitcoinS]
{
	\label{fig::rate_BitcoinM}
	\includegraphics[width=0.4\linewidth]{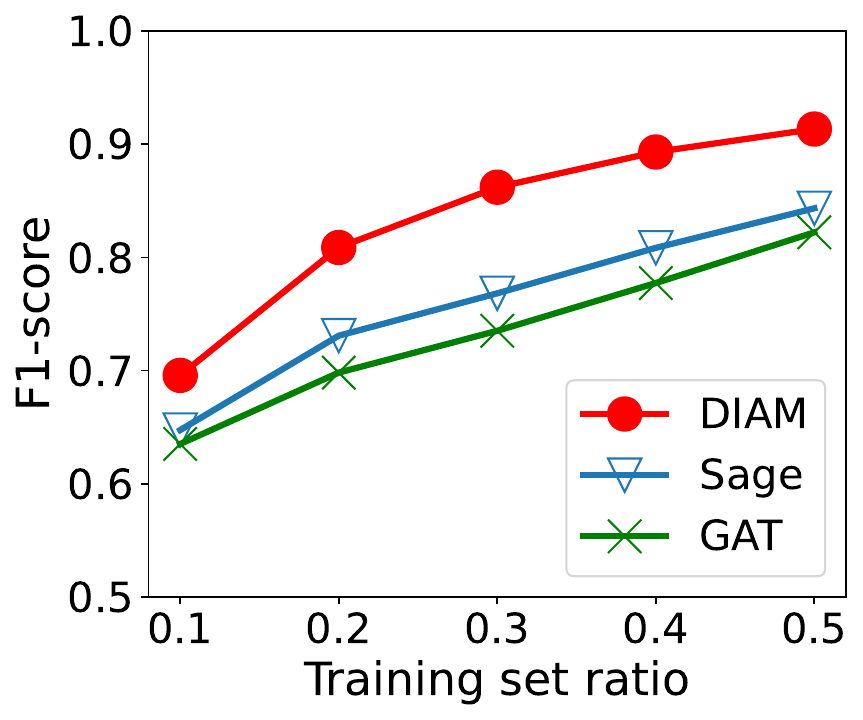} 
}
\subfigure[\bitcoinL]
{
	\label{fig::rate_BitcoinL}
	\includegraphics[width=0.42\linewidth]{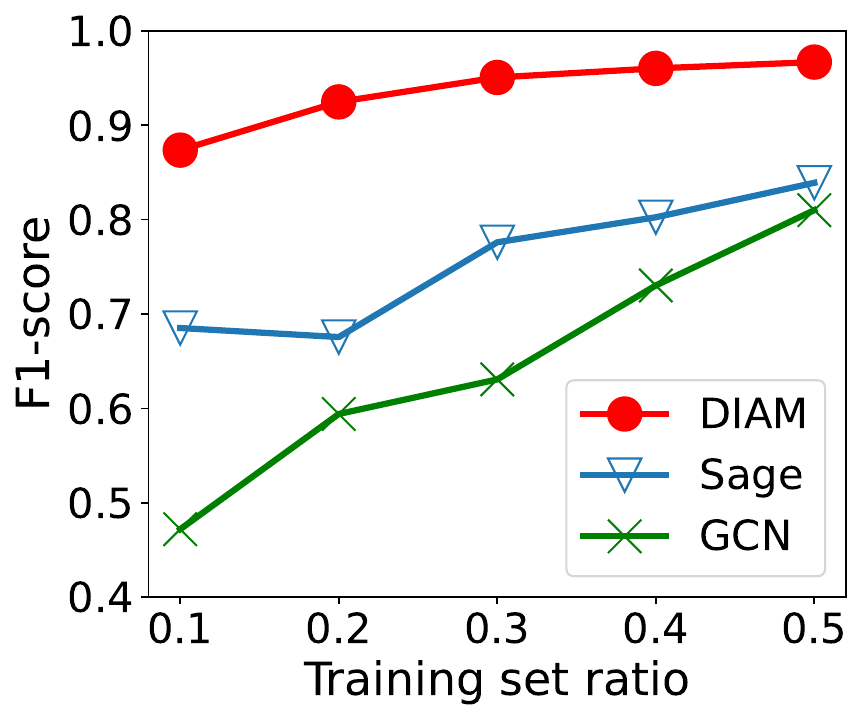}  
}
\vspace{-2mm}
\caption{Varying training set size ratio}
\label{fig:rate}
\vspace{-1mm}
\end{figure}

\header
\textbf{Varying  training data ratio.} To compare the performance of \algo with baselines under the situation with insufficient training data, we vary the percentage of training data from $10\%$ to $50\%$. The F1 results on all datasets are reported in Figure \ref{fig:rate}, where \algo and the top-2 best baselines per dataset are evaluated. 
The overall observation is that the F1 scores of all methods decrease as the amount of training data decrease; meanwhile, \algo keeps achieving the highest effectiveness. 
On \ethsecond in Figure \ref{fig::rate_EthereumS}, we compare \algo with the top-2 baselines \sage and \gcn of the dataset (see Table \ref{tab:overall}).
 For different sizes of training data, \algo keeps outperforming the baselines. 
 Similarly, we compare \algo with \sage and \edgeprop on \ethphish in Figure \ref{fig::rate_EthereumP},   compare \algo with \sage and \gat on \bitcoinS in Figure \ref{fig::rate_BitcoinM}, compare \algo with \sage and \gcn on \bitcoinL
in in Figure \ref{fig::rate_BitcoinL}.
 The results in all figures show that \algo consistently achieves the highest F1 scores, regardless of the volume of training data.
Another observation   is that  the performance of \algo is relatively stable on the largest \bitcoinL. Compared to training with 50\% of the data, training with 10\% of the data only resulted in a 9.6\% decrease in model performance. While the two other competitors decreased 18.3\% (\sage) and 41.7\% (\gcn), respectively, which validates the capability of \algo to leverage abundant data to obtain expressive representations for illicit account detection.

%% file: conclusion.tex
\section{Conclusion}

We present \algo, an effective discrepancy-aware multigraph  neural network for the problem of illicit account detection on cryptocurrency transaction networks. 
The core techniques in \algo include \edgeseq that leverages sequence models to automatically learn node representations capturing both incoming and outgoing transaction patterns, and a new Multigraph Discrepancy module MGD, which is able to learn high-quality representations to distinguish the discrepancies between illicit and normal nodes.
We conduct extensive experiments on 4 large cryptocurrency datasets, and compare \algo against 15 existing solutions. The comprehensive experimental results show that \algo consistently achieves superior performance. 
Note that the multigraph model in this paper can also describe other transaction networks besides cryptocurrencies, such as online payment data by tech firms, \eg AliPay and PayPal. Hence, in the future, in addition to cryptocurrency transaction networks, we plan to  apply our method to other types of transaction networks to further validate its effectiveness.